\pdfoutput=1

\documentclass[11pt]{article}

\usepackage[final]{acl}

\usepackage{times}
\usepackage{latexsym}

\usepackage[T1]{fontenc}

\usepackage[utf8]{inputenc}

\usepackage{microtype}

\usepackage{inconsolata}

\usepackage{makecell}
\usepackage{comment}
\usepackage{dsfont}
\usepackage{amsmath}
\usepackage{mathtools}
\usepackage{graphicx,multirow,makecell}
\usepackage{booktabs, tabularx}
\usepackage{xcolor,colortbl}
\usepackage{color,soul}
\usepackage{subcaption}
\usepackage{fixltx2e}
\usepackage[shortlabels]{enumitem}
\usepackage[symbol]{footmisc}
\renewcommand*{\thefootnote}{\arabic{footnote}}

\newcommand\ie{\emph{i.e.}}

\definecolor{as}{HTML}{0000FF} 
\definecolor{op}{HTML}{FF8000} 
\definecolor{sp}{HTML}{009900} 
\definecolor{ac}{HTML}{CC0000} 

\title{It is Simple Sometimes: A Study On Improving Aspect-Based Sentiment Analysis Performance}

\author{Laura Cabello\thanks{\hspace{2mm} Work done during internship at Sony AI.}\\ Department of Computer Science,\\ University of Copenhagen \\
  \texttt{lcp@di.ku.dk} \\\And
  Uchenna Akujuobi \\
  Sony AI, Japan \\
  \texttt{uchenna.akujuobi@sony.com} \\}

\begin{document}
\maketitle
\begin{abstract}

Aspect-Based Sentiment Analysis (ABSA) involves extracting opinions from textual data about specific entities and their corresponding aspects through various complementary subtasks. 
Several prior research has focused on developing \textit{ad hoc} designs of varying complexities for these subtasks.
In this paper, we present a generative framework extensible to any ABSA subtask. 
We build upon the instruction tuned model proposed by \citet{scaria2023instructabsa}, who present an instruction-based model with task descriptions followed by in-context examples on ABSA subtasks. 
We propose PFInstruct, an extension to this instruction learning paradigm by appending an NLP-related task prefix to the task description. 
This simple approach leads to improved performance across all tested SemEval subtasks, surpassing previous state-of-the-art (SOTA) on the ATE subtask (Rest14) by +3.28 F1-score, 
and on the AOOE subtask by an average of +5.43 F1-score across SemEval datasets.
Furthermore, we explore the impact of the prefix-enhanced prompt quality
on the ABSA subtasks and find that even a noisy prefix enhances model performance compared to the baseline. 
Our method also achieves competitive results on a biomedical domain dataset (ERSA).
\end{abstract}

\section{Introduction}
\begin{figure}
    \centering
    \includegraphics[width=\columnwidth, clip]{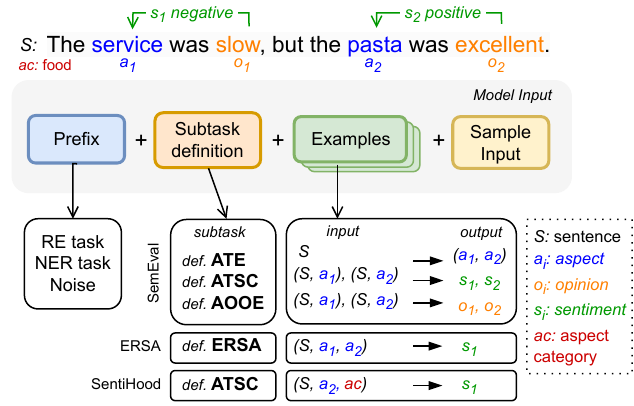}
    \caption{Illustration of model input and ABSA subtasks examined in this paper. The \textbf{prefix} can vary between NLP-related tasks (instruction) or textual noise (random words), followed by the \textbf{subtask definition}, few \textbf{examples} and the corresponding \textbf{sample input} for each subtask. The model is expected to follow the instructions and generate a prediction. Subtasks belong to three distinct data sources: SemEval, ERSA and SentiHood from different domains. }
    \label{fig:overview}
\end{figure}

User-generated reviews on e-commerce and social media platforms benefit both consumers and stakeholders. With the exponential growth of data, developing reliable tools for 
understanding the sentiment of online review texts
is essential to moderate online content, enable effective decision-making and customer satisfaction. 
\citet{Liu_2012} proposed Aspect-Based Sentiment Analysis (ABSA) as a step towards fine-grained sentiment analysis of specific aspects.
ABSA involves the detection of opinions (\textit{\textcolor{op}o}) and sentiment (\textit{\textcolor{sp}s}) associated with particular aspects (\textit{\textcolor{as}a}) in a text (\textit{S}). Figure~\ref{fig:overview} summarizes the five ABSA subtasks considered in this paper. 

Instruction-based learning has emerged as a promising paradigm to successfully tune large language models (LLM) on a variety of tasks \cite{wei2022finetuned,wang-etal-2022-super,singhal2022large,gupta2023instruction}.
The attraction of instruction-based learning is the ability to steer the base model behaviour to follow instructions 
\cite{ouyang2022training, bowman2023things}.
In the context of ABSA, \citet{scaria2023instructabsa} proposed InstructABSA,
an instruction-based model based on a 200M-parameter Tk-Instruct model \cite{wang-etal-2022-super}. 
Their best performing setting, dubbed as InstructABSA2, frames the task instruction as a task definition followed by two positive, negative, and neutral examples. We build our experiments upon the same setting.

In this paper, we propose PFInstruct, an extension to the InstructABSA framework with the introduction of prefix prompt. Specifically, 
we append a prefix to the task definition, extend the evaluation to domains like biomedicine and urban neighbourhoods, and formulate \textit{all} the subtasks as a generative task. 
The prefix aims at instructing the model on a related NLP task, namely Relation Extraction (RE) or Named Entity Recognition (NER), so the target text \textit{S} is seen on a different task context. We postulate this approach helps to collect richer semantic information about the main entities in \textit{S}, which allows the model to make a more informed prediction about the ABSA subtasks.
We also consider to use a randomly generated (noise) prefix. We observe that not only it boosts --to a lesser degree-- average performance, but it also makes the model more robust to out-of-domain data. 

\paragraph{Contributions} 
We introduce a simple approach to solve ABSA subtasks, that is, a prefix prompt followed by instructions. Our approach outperforms previous SOTA on several tasks despite being based on a 200M model. We conduct extensive analysis on five subtasks from different domains: customer reviews from SemEval 2014, 15 and 16 \cite{pontiki-etal-2014-semeval,pontiki-etal-2015-semeval,pontiki-etal-2016-semeval}, on a biomedical domain dataset, ERSA \cite{cerm}, and on user comments from SentiHood \cite{saeidi-etal-2016-sentihood}. We assess the effect of the prefix prompt in terms of prompt quality (RE, NER or noise) and domain generalization (out-of-domain performance). 

We make our code publicly available to ensure reproducibility and foster future research.\footnote{\url{https://github.com/lautel/PFInstruct}}

\section{Related Work}
Current state-of-the-art (SOTA) solutions on SemEval datasets include LSA \cite{yang2023improving} on ATSC, which leverage the use of gradient descent to design a differential-weighted approach, BARTABSA \cite{yan-etal-2021-unified} on AOOE, based on an end-to-end generative framework, and \citet{sun-etal-2019-utilizing} on SentiHood (ATSC).
On a similar line of research to \citet{scaria2023instructabsa} and our work, \citet{varia-etal-2023-instruction} propose IT-MTL, multi-task prompting with instructional prompts to tackle ABSA subtasks as a question-answering problem.
For a review of popular ABSA datasets, we refer to \citet{chebolu2023survey}.

To compare to related work \cite{yan-etal-2021-unified,li-etal-2021-dual-graph,mao2021joint,varia-etal-2023-instruction,yang2023improving}, we report macro-averaged F1-score.

\section{Background and Methodology}
\subsection{Background: ABSA}\label{sec:absa}
ABSA subtasks can be classified into single output and compound output subtasks. In single output subtasks such as ATE (Aspect-Term Extraction), ATSC (Aspect-Term Sentiment Classification), and AOOE (Aspect-Opinion Extraction), the output is limited to either the aspect \textit{\textcolor{as}a}, opinion \textit{\textcolor{op}o} or sentiment \textit{\textcolor{sp}s} from a given text \textit{S}. Compound output subtasks ask for a combination of the $\{a, o, s \}$ entity types. Many researchers \cite{xue-li-2018-aspect,wu2020latent,pouran-ben-veyseh-etal-2020-introducing,yang2023improving} focus only on the former. For convenience, we do the same.\footnote{The application of our method to compound-output subtasks is straightforward.} 

In addition, we evaluate our method on ERSA \cite{cerm} and SentiHood (ATSC) \cite{saeidi-etal-2016-sentihood} tasks. 
ERSA is an extension of the ABSA task, where, given a text $S$ and two aspects (or entities) \textit{\textcolor{as}{a$_1$}} and \textit{\textcolor{as}{a$_2$}}, where \textit{\textcolor{as}{a$_1$}} $\neq$ \textit{\textcolor{as}{a$_2$}}, the goal is to determine the sentiment polarity \textit{\textcolor{sp}s} of the relationship between the aspects given $S$. ERSA targets biomedical texts but it does not require to have factual knowledge about the entities. 
SentiHood, defined as an extension of ATSC, requires to classify the sentiment \textit{\textcolor{sp}s} towards each aspect \textit{\textcolor{as}{a$_i$}} of one or more aspect categories \textit{\textcolor{ac}{ac}}.
Appendix~\ref{app:experimental} describes the datasets in more detail.


\subsection{Methodology}
We propose an extension of the InstructABSA framework.
Given a sample \textit{S}, we construct a prompt that consists of 4 components which we detail below.

\begin{enumerate}[align=left,leftmargin=*,wide = 0pt,topsep=0pt,itemsep=-1ex,label={\roman*})]
    \item[] \textbf{Prefix}. An initial instruction to explicitly ask the model to solve an NLP task on the sample \textit{S}. The purpose of this prefix is to involve the main entities in \textit{S} in an preliminary NLP task, which can inform the subsequent ABSA subtask on the main entities in \textit{S} to determine the correct output. 
    This NLP task can be Named Entity Recognition (NER) or Relation Extraction (RE). RE is applied if the aspects \textit{\textcolor{as}{a$_i$}} are part of the task input and the sample contains at least two entities (or aspects).
    We also analyse the effect of having a noisy prefix prompt composed of random words. 
    \item[] \textbf{Task definition}. A succinct overview of the ABSA subtask. In sentiment classification tasks, we also include the set of pre-defined classes.
    \item[] \textbf{Examples}. A set of two positive, negative, and neutral in-domain examples.
    \citet{scaria2023instructabsa} carry out an extensive analysis on the effect of different task definitions and example manipulations. As our method extends their approach, we fix the task definition and set of examples to match their best performing set-up, namely InstructABSA2.
    \item[] \textbf{Sample input}. Similar to the in-context examples, we provide the model the input \textit{S} and expect the model will follow the instructions and generate the corresponding output.
\end{enumerate}

At inference time, we repeat the same structure with sample inputs from the test split. 
Appendix~\ref{app:prompt} shows specific examples of the final prompts.

\section{Results}
\label{sec:results}
We present the main results of our experimental set-up. report macro-averaged F1-score averaged across five random initialization seeds and the standard deviation. Details about fine-tuning settings are provided in Appendix~\ref{app:experimental}.

\subsection{Analysis of SemEval subtasks}\label{sec:semeval}
Tables~\ref{tab:ate}--\ref{tab:aooe} show results of ATE, ATSC\footnote{To maintain consistency with existing methods, we also remove instances labelled as `conflict' \cite{chen-etal-2017-recurrent,li-etal-2021-dual-graph,scaria2023instructabsa}.} and AOOE subtasks respectively. 
Our method achieves superior performance (F1-scores) when compared with previous SOTA methods across all subtasks. Specifically,
we observe that setting an NLP-related task prefix outperforms previous models in 6/12 cases. Interestingly, adding a noise prefix surpasses previous approaches in 4 of the remaining 6 cases. 
These results validate our initial hypothesis: instructing the model to solve a related NLP task for the target text \textit{S} seems to complement the model's understanding of the main entities in \textit{S}, which leads to more accurate predictions.\footnote{We look into the subset of predictions for input samples with two or more aspects in ATSC and AOOE tasks and observe a similar trend to what is reported here.}

In general, providing a random prefix improves model performance compared to not including a prefix at all (see results from InstructABSA2 rows) in the three subtasks. This is more pronounced in ATE. Contrary to ATSC and AOOE, ATE requires the model to make a prediction based solely in the input text \textit{S}, \ie, it does not include a target aspect \textit{\textcolor{as}{a}} as input. 
This setting causes the effect of focusing on entity recognition or having random prefixes to be similar on average across datasets: $\text{F1}=90.35$ (PFInstruct-NER) compared to $\text{F1}=90.53$ (PFInstruct-Noise).
However, the disparate variance in PFInstruct-Noise makes PFInstruct-NER an overall better model choice.

\begin{table}[ht!]
\centering
\resizebox{\columnwidth}{!}{
\begin{tabular}{lcccc|c}
\toprule
Model &Lapt14 &Rest14 &Rest15 &Rest16 &Avg.\\ \midrule
BARTABSA$^\dagger$ & 83.52 &87.07 &75.48 &- & -\\
InstructABSA2$^\dagger$ & 92.30 &92.10 &76.64 &80.32 &85.34 \\
\midrule
PFInstruct-NER &92.65$\pm$0.70 &\textbf{95.38}$\pm$0.10 &82.86$\pm$1.15 &90.51$\pm$0.72 &90.35$\pm$0.67 \\
PFInstruct-Noise &\textbf{92.90}$\pm$3.95 &94.92$\pm$0.46 &\textbf{83.58}$\pm$0.61 &\textbf{90.73}$\pm$2.41 &90.53$\pm$1.86 \\
\bottomrule
\end{tabular}
}
\caption{
F1-scores for ATE subtask. Avg stands for average across datasets. $^\dagger$Results from original papers.
}
\label{tab:ate}
\end{table}

\begin{table}[ht!]
\centering
\resizebox{\columnwidth}{!}{
\begin{tabular}{lcccc|c}
\toprule
Model &Lapt14 &Rest14 &Rest15 &Rest16 &Avg.\\ \midrule
LSA$_T$-X$^\dagger$ &\textbf{83.93} &86.26 &- &- &-\\
Dual-MRC$^\dagger$ &75.97 &82.04 &73.59 &- &-\\
BARTABSA$^\dagger$ &76.76 &75.56 &73.91 &- &-\\
InstructABSA2$^\ddagger$&81.66$\pm$0.80 &86.70$\pm$0.63 &85.06$\pm$1.05 &\textbf{93.01}$\pm$0.71 &86.61$\pm$0.80
\\
\midrule
PFInstruct-RE &82.57$\pm$1.05 &86.68$\pm$0.71 &\textbf{86.16}$\pm$1.17 &92.51$\pm$0.20 &86.98$\pm$0.78\\
PFInstruct-NER &81.63$\pm$0.28 &86.66$\pm$0.92 &85.61$\pm$1.90 &86.60$\pm$0.35 &85.13$\pm$0.86 \\
PFInstruct-Noise &80.88$\pm$1.02 &\textbf{86.88}$\pm$1.50 &84.32$\pm$0.59 &91.54$\pm$0.51 &85.91$\pm$0.91
\\\bottomrule
\end{tabular}
}
\caption{F1-scores for ATSC subtask. Avg stands for average across datasets. $^\dagger$Results from original papers. $^\ddagger$Results are reproduced by us, since \citet{scaria2023instructabsa} report accuracy.}
\label{tab:atsc}
\end{table}

\begin{table}[ht!]
\centering
\resizebox{\columnwidth}{!}{
\begin{tabular}{lcccc|c}
\toprule
Model &Lapt14 &Rest14 &Rest15 &Rest16 &Avg.\\ \midrule
Dual-MRC$^\dagger$ &79.90 &83.73 &74.50 &83.33 &80.37 \\
BARTABSA$^\dagger$ &80.55 &85.38 &80.52 &87.92 &83.59 \\
InstructABSA2$^\dagger$ &77.16 &81.08 &81.34 &83.27 &80.71 \\ \midrule
PFInstruct-RE &\textbf{84.04}$\pm$0.31 &90.10$\pm$0.36 &\textbf{89.56}$\pm$0.55 &88.51$\pm$0.38 &88.05$\pm$0.40 \\
PFInstruct-NER &83.43$\pm$1.61 &\textbf{91.47}$\pm$0.33 &89.11$\pm$0.28 &\textbf{92.08}$\pm$0.81 &89.02$\pm$0.65 \\
PFInstruct-Noise &81.06$\pm$1.17 &91.00$\pm$0.55 &87.56$\pm$0.72 &90.70$\pm$0.31 &87.58$\pm$0.68
\\\bottomrule
\end{tabular}
}
\caption{F1-scores for AOOE subtask. Avg stands for average across datasets. $^\dagger$Results from original papers.}
\label{tab:aooe}
\end{table}

We postulate that the additional prefix (random or NLP related) enhances the model's ability to selectively filter out irrelevant information to the final task, thereby bolstering its resilience to textual inaccuracies such as misspellings or grammatical errors. However, including a random prefix has the negative side effect of making the model more sensitive to its initial random weights, as shown by the higher variance of PFInstruct-Noise across settings.


\paragraph{Error Analysis}
To gain a better insight on the benefits of introducing a random prefix (PFInstruct-Noise) compared to not introducing a prefix at all, we perform a case study on the incorrectly classified examples by PFInstruct without prefix, i.e., we reproduce results with the same settings from InstructABSA2.
From these errors, 23\% are correct in ATE with PFInstruct-Noise and 28\% in AOOE.
In sentiment classification (ATSC), both setups missclassify the same samples. 
Focusing on ATE, Table \ref{tab:ateanalysis} showcases examples where 
the introduction of a noise prefix seems beneficial for
a better understanding of misspelling errors and especial jerga. 
We also find cases where aspect term extraction with PFInstruct-Noise is more comprehensive, including descriptive adjectives (rows i and iii).


\begin{table}[]
\centering
\resizebox{\columnwidth}{!}{
\begin{tabular}{llll} 
\toprule
~ & Sample & Prefix & Output  \\\midrule
\multirow{3}{*}{i} & \multirowcell{3}[0pt][l]{i highly recommend this place \\to all that want to try \textcolor{as}{indain food} \\for the first time.} & No prefix & place  \\
 && Noise & indain food \\\\\midrule
\multirow{3}{*}{ii} & \multirowcell{3}[0pt][l]{\textcolor{as}{Screen} - although some people \\might complain about low \textcolor{as}{res} \\which I think is ridiculous.} & No prefix & screen  \\
 && Noise & screen, res \\\\\midrule
\multirow{3}{*}{iii} & \multirowcell{3}[0pt][l]{Really Lovely \textcolor{as}{dining experience}\\ in the midst of buzzing\\ midtown area.} & No prefix & dining  \\
 && Noise & dining experience \\\\\midrule
\multirow{3}{*}{iv} & \multirowcell{3}[0pt][l]{They have \textcolor{as}{homemade pastas}\\ of all kinds -- I recommend the \\\textcolor{as}{gnocchi} -- yum!} & No prefix & pastas, gnocchi  \\
 && Noise & homemade pastas, \\&&&gnocchi  
\\\bottomrule
\end{tabular}}
\caption{Case study on the ATE subtask on examples where the model fails in the absence of a prefix, but PFInstruct-Noise outputs the correct target aspect(s). We observe that PFInstruct-Noise is more robust to misspellings errors and chatspeak (i, ii) and extracts more detailed answers (iii, iv).}
\label{tab:ateanalysis}
\end{table}

Other work in NLP also find beneficial the addition of noise. Amongst others, \citet{jain2024neftune} show that noisy embeddings improve instruction fine-tuning, and \citet{cuconasu2024power} prove that including irrelevant documents can enhance performance of retrieval-augmented generation (RAG) systems. 

\subsection{Analysis of ERSA and SentiHood}
Table~\ref{tab:ersa} shows results on both datasets\footnote{To obtain comparable results to existing methods \cite{cerm, saeidi-etal-2016-sentihood}, we utilize only the four most frequent aspects in SentiHood.}, where we can see that ERSA is the clear exception to the trend observed in Section~\ref{sec:semeval}: 
the choice of a prefix is important as it can negatively affect model performance.

The inherent nature of ERSA presents a more significant challenge than other ABSA subtasks, since the sentiment expressed in a text \textit{S} may not necessarily reflect the sentiment of the relationship between the target entities, \textit{\textcolor{as}{a$_1$}} and \textit{\textcolor{as}{a$_2$}} \cite{cerm}. 
In this case, in-context noise hurts model performance the most. The model needs to adapt to a specialised domain and learn the nuances of the task. 
In terms of NLP-task prefix, leveraging the knowledge of \textit{\textcolor{as}{a$_1$}} and \textit{\textcolor{as}{a$_2$}} to reason about their semantic relationship (PFInstruct-RE) improves model performance over general entity recognition (PFInstruct-NER). However, it does not surpass the model performance achieved without prompt prefixes (InstructABSA2 setup). 

\paragraph{Error Analysis}
We examine the misclassified examples by PFInstruct-RE to better understand why, contrary to what we observed in Section~\ref{sec:semeval}, the absence of prefixes appears to be beneficial (see InstructABSA2 in Table~\ref{tab:ersa}).
We observe that in $\sim50\%$ of these cases, annotators have labelled the sentiment based on the \emph{meaning of the full sentence} instead of focusing on the relationship between the given entities \emph{in} the context. For instance, \textit{`Treatment with ERY resulted in fewer inflammatory cells and \textcolor{as}{\underline{cytokines}} in the BALF, and fewer \textcolor{as}{\underline{emphysema}}-associated changes...'} is labelled as \textcolor{sp}{\emph{negative}} despite the relationship between the target entities being \textcolor{sp}{\emph{neutral/none}}. 

\renewcommand*{\thefootnote}{\fnsymbol{footnote}}

\begin{table}[]
\centering
\resizebox{\columnwidth}{!}{
\begin{tabular}{lcc}
\toprule
Model & ERSA & SentiHood (ATSC) \\ \midrule
CERM\footnotemark[2] &71.0 &88.50 \\
BERT-pair-QA-M\footnotemark[2] & - &93.60 \\
InstructABSA2 &\textbf{70.76}$\pm$0.41 &94.90$\pm$0.07\\
\midrule
PFInstruct-RE &70.31$\pm$0.14 &- \\
PFInstruct-NER &70.00$\pm$0.50 &93.83$\pm$0.03 \\
PFInstruct-Noise &64.72$\pm$0.59 &\textbf{95.11}$\pm$0.02
\\\bottomrule
\end{tabular}
}
\caption{F1-scores for ERSA and SentiHood tasks. Results from InstructABSA2 are reproduced by us.}
\label{tab:ersa}
\end{table}
\footnotetext[2]{Results from the original papers (acc).}
\renewcommand*{\thefootnote}{\arabic{footnote}}

\subsection{Domain generalization}\label{sec:ood}
\begin{figure}
\centering
    \begin{subfigure}{0.24\textwidth}
    \includegraphics[width=\textwidth]{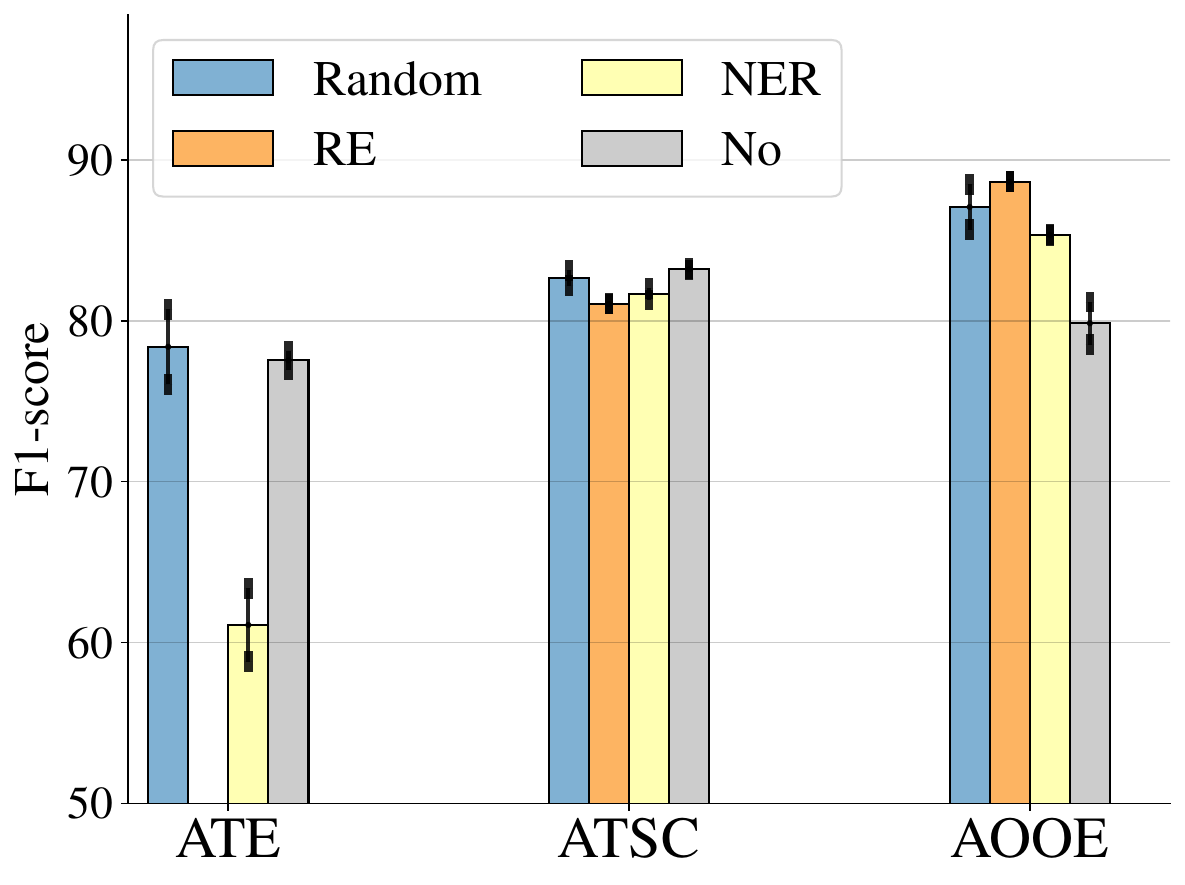}
    \caption{tr: Lapt14, tst: Rest14}
    \end{subfigure}%
    \begin{subfigure}{0.24\textwidth}
    \includegraphics[width=\textwidth]{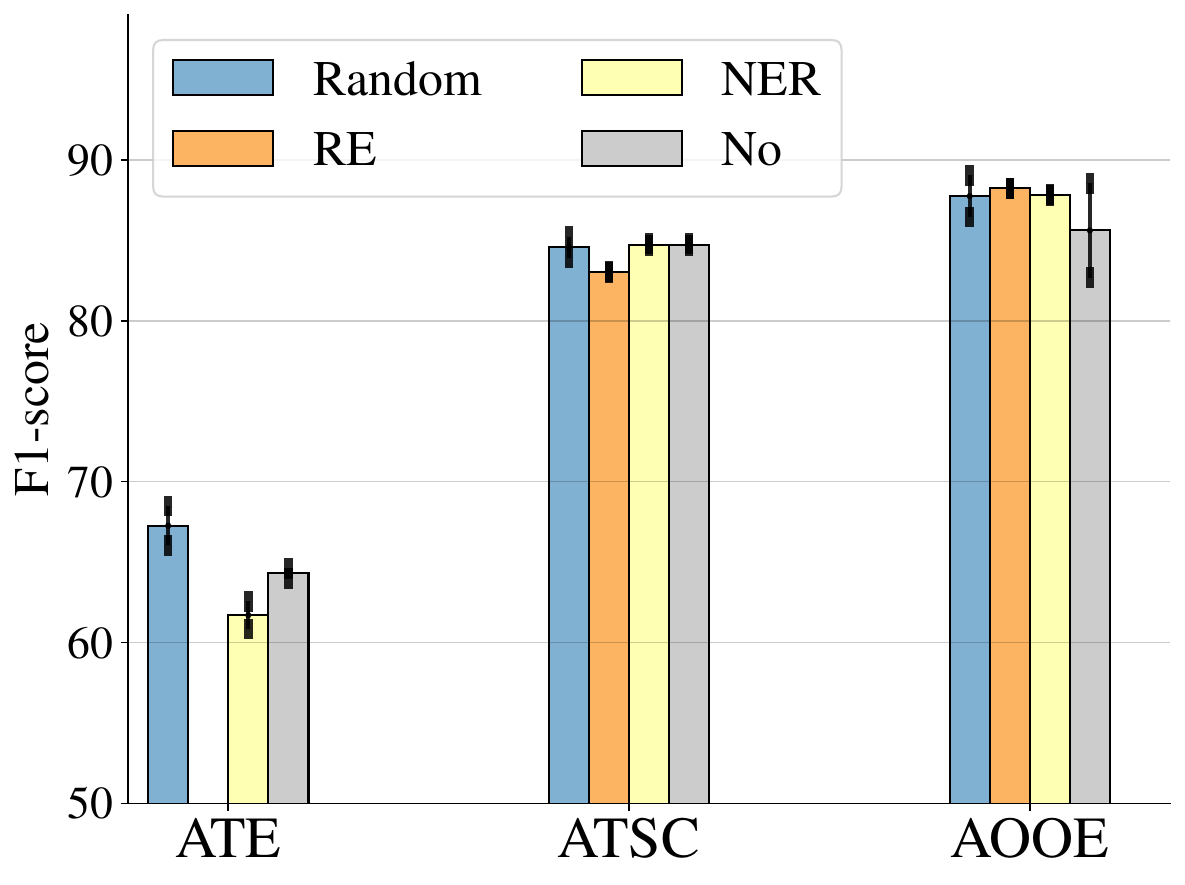}
    \caption{tr: Lapt14, tst: Rest15}%
    \end{subfigure}
    \begin{subfigure}{0.24\textwidth}
    \includegraphics[width=\textwidth]{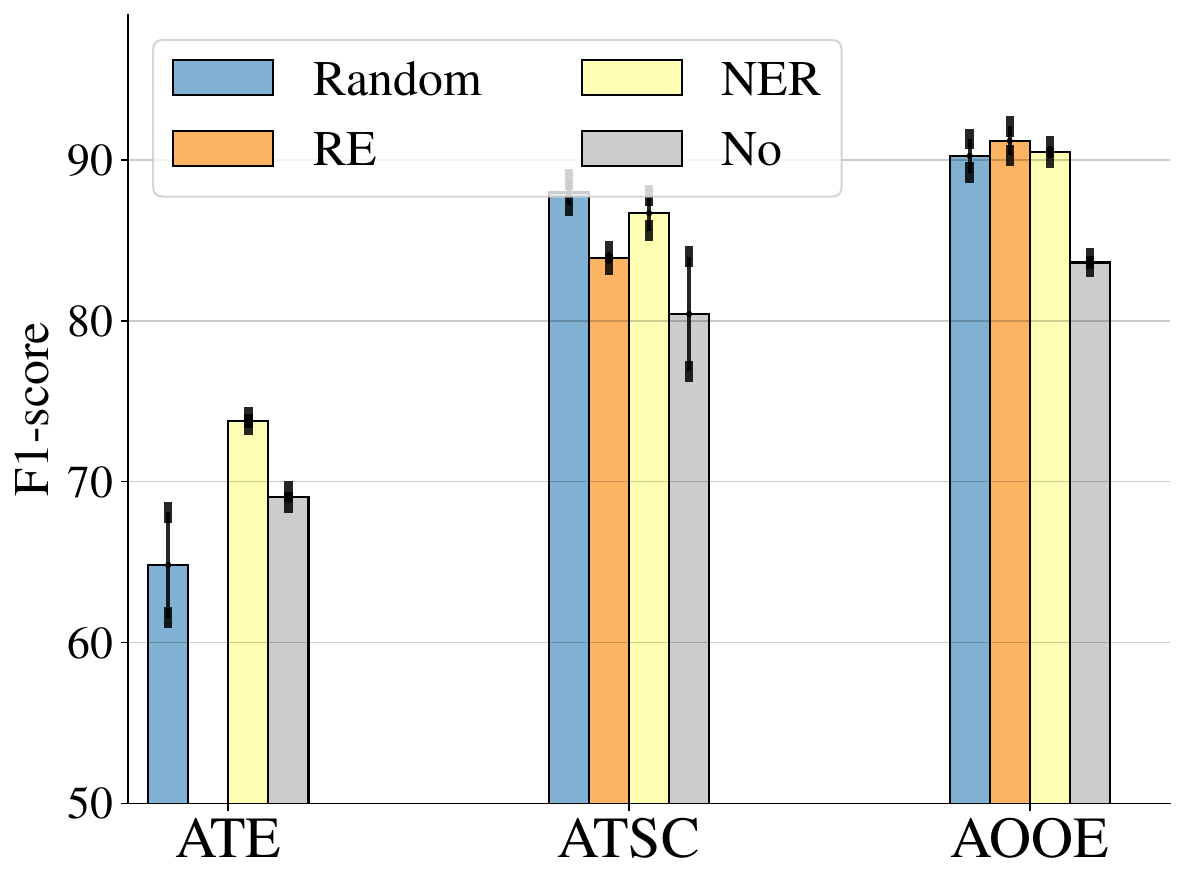}
    \caption{tr: Lapt14, tst: Rest16}
    \end{subfigure}
    \begin{subfigure}{0.23\textwidth}
    \includegraphics[width=\textwidth]{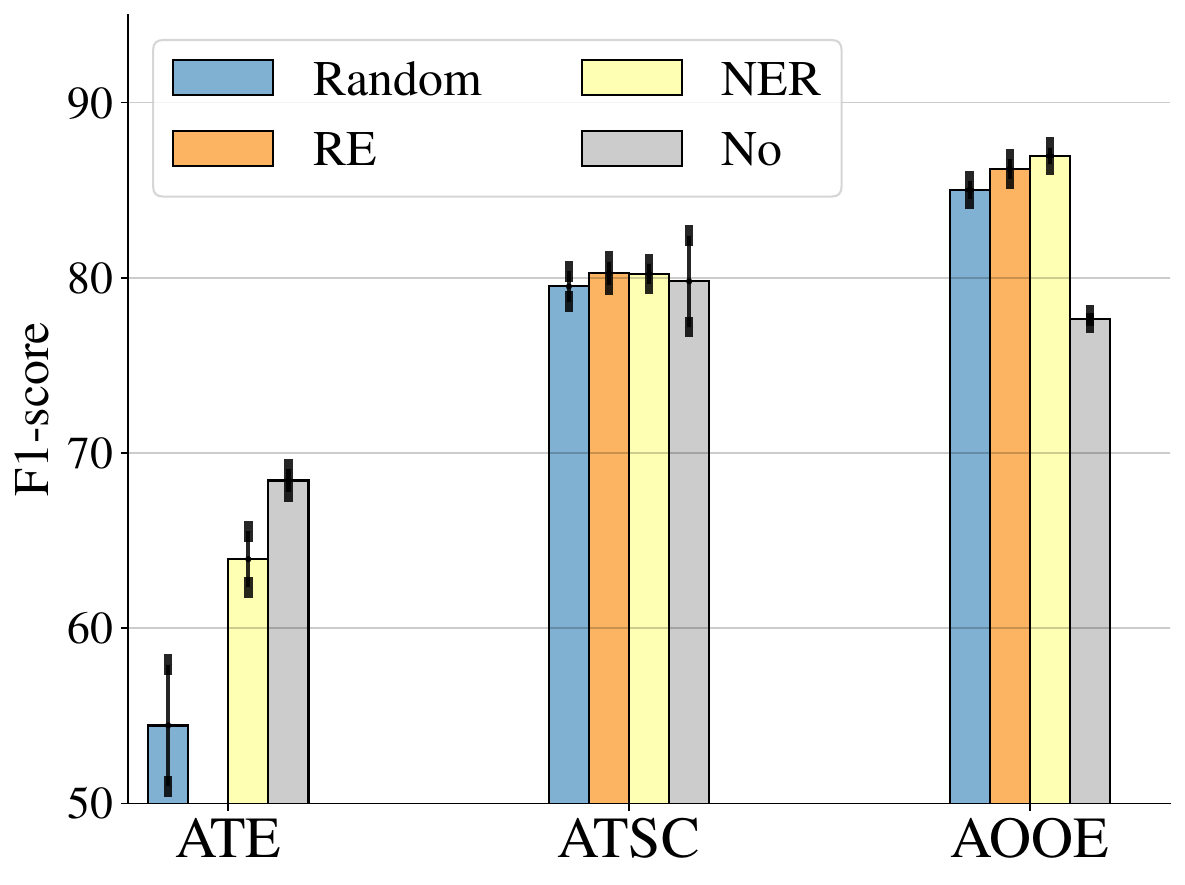}
    \caption{tr: Rest*, tst: Lapt14}%
    \end{subfigure}
\caption{Out-of-domain evaluation. F1-scores are averaged across five random initialization seeds; error bars show the standard deviation.
Models are trained (`tr') on one domain and evaluated (`tst') on a distinct domain.
Legends indicate the prefix prompt used. `No' stands for no use of prefix. RE is not evaluated for ATE (see Section~\ref{sec:results}).}\label{fig:ood}
\end{figure}

Results from Tables~\ref{tab:ate}--\ref{tab:ersa} demonstrate the viability of our method with in-domain data in SemEval and SentiHood, while remaining competitive in ERSA. 
In this section, we explore the robustness of our models when evaluated on out-of-domain data in SemEval. 
Figure~\ref{fig:ood} shows results when training the models on laptops domain (Lapt14) and evaluating on restaurants domain (Rest14, Rest15, Rest16), and vice versa.

As expected, we observe a general drop in performance compared to training in-domain, which is especially large in ATE. However, when training in restaurant domain and evaluating in Lapt14 for AOOE --see Figure~\ref{fig:ood} (d)--, all model variants surpass their respective in-domain results with a major improvement of $+3.53$ F1 in PFInstruct-NER. We conclude that the addition of more training data is beneficial for this task.


The addition of any kind of prefix helps to make the model more robust to out-of-domain data for AOOE, while it does not significantly hurt performance for ATSC. Interestingly in this task, the large variance shown by the model without prefix is reduced by the addition of a prefix, especially with RE and NER prefixes --see Figure~\ref{fig:ood} (c), (d)--.


While the strategy of adding a noisy prefix seem beneficial to out-of-domain data performance too, looking closer we make two observations: \emph{i)} PFInstruct-Noise models show large variance regardless of the domain, and \emph{ii)} the drop in performance when evaluated on out-of-domain data (compared to in-domain) is larger for PFInstruct-Noise models. Therefore, these results suggest that an NLP task prefix makes the model more robust to domain shifts.


\section{Discussion}
As shown in Section \ref{sec:results}, the addition of an NLP task prefix can boost model performance, especially in ATE and AOOE. While the addition of a noise prefix also seems beneficial in most of the cases tested, it also comes with high fluctuations in performance depending on the random initialisation of the model’s weights. For this reason, we would caution against this choice for a final application. However, the a priori effectiveness of PFInstruct-Noise echoes the question posed by \citet{kung-peng-2023-models}, do models really learn to follow instructions? Our analysis provides evidence that performance gains in ABSA subtasks can come from seeing the target entities (or aspects) in a preliminary NLP task instruction, suggesting the utility of this instruction. Nevertheless, our results with PFInstruct-Noise also highlights the need for more in depth analysis of instruction based learning and evaluation.

\section{Conclusion}
In this paper, we present PFInstruct, a simple yet effective prefix prompting strategy to instruction fine-tune a language model on ABSA subtasks. 
We analyse the impact of the prefix prompt's quality on in-domain and out-of-domain data and observe that even a random prefix improves average model performance compared to the InstructABSA baseline.
We evaluate our method on domains such as customer reviews, biomedical text and user comments, and show that it outperforms previous SOTA approaches on most of the tasks tested and achieves competitive performance (F1-score) in the rest.

\section*{Limitations}
Our study builds upon an instruction tuned language model, Tk-Intruct, and therefore it inherits its limitations. However, we have done an extensive analysis on a variety of domains and task settings, namely SemEval (customer reviews of laptops and restaurants) ERSA (healthcare) and SentiHood (user comments about urban neighbourhoods), proving the generalizability of our method. 
Despite our method can be easily applied to other language models, the effectiveness of PFInstruct have not been tested with other model architectures nor model sizes. 

Our approach reduces the effective input sequence length of the model, since we need to allocate input tokens for the prefix prompt. While this side-effect is worth noting, it has not supposed an issue for the current experiments (maximum sequence length for Tk-Instruct: $512$ tokens, average prompt length excluding input sentences: $348$ tokens with Relation Extraction or random (noise) prompt, 304 tokens with Named Entity Recognition prompt).
In addition, our work is limited to an English language model and English texts. Future studies should prove the validity of our approach in languages other than English.

\section*{Ethics Statement}
The models and datasets used in this study are publicly available, and we strictly follow their terms of use. We meet the ethical implications of previous research related to the data sources. It is important to acknowledge the presence of inherent biases to the data and models used in this study, but we do not anticipate other ethical risks derived from our work. 


\bibliography{custom}

\clearpage
\appendix

\section{Experimental details}
\label{app:experimental}

\subsection{Data}


Distribution of datasets used:

\begin{enumerate}[itemsep=0ex]
    \item[] \textbf{Lapt14}: 3045 Train, 800 Test
    \item[] \textbf{Rest14}: 3041 Train, 800 Test
    \item[] \textbf{Rest15}: 1315 Train, 685 Test
    \item[] \textbf{Rest16}: 2000 Train, 676 Test
    \item[] \textbf{Hotel15}: 266 Test
    \item[] \textbf{ERSA}: 8183 Train, 909 Validation, 2274 Test
    \item[] \textbf{SentiHood}: 5215 Train (2460 from top-4 aspect categories), 610 Validation, 1216 Test 
\end{enumerate}


\subsection{Experiments} 

We instruction fine-tune the model checkpoint \texttt{Tk-Instruct-base-def-pos} with the following hyprparameters:

\begin{itemize}[itemsep=0ex]
    \item N. epochs: 4%
    \item Batch Size: 16 for ATE, ATSC, ERSA; 8 for AOOE and SentiHood. Batch sizes explored: \{8, 16\}%
    \item Learning rate: 1e-4 for ATE, ATSC, AOOE and SentiHood; 5e-5 for ERSA. Learning rates explored: \{1e-5, 5e-5, 1e-4, 5e-4\}%
    \item Warmup ratio: 0.1%
    \item Regularization: weight decay, 0.01%
\end{itemize}

Experiments were performed in 2 A10G GPUs.
Hyperparameter tuning was performed based on validation performance in each dataset. If a validation split was not originally provided, we held out 10\% of the train split.


\section{Prompt Examples}
\label{app:prompt}

Table~\ref{tab:prefix-example} and Table~\ref{tab:prefix-example2} provide examples of prefixes for two given input texts \textit{S}. Each table illustrate the three prefix types defined in the paper. We set the Noise-prefix length to 50 words to match the average length of the RE-prefix.

\begin{table}[]
    \centering
    \resizebox{\columnwidth}{!}{
    \begin{tabular}{p{0.2\linewidth} p{0.75\linewidth}}
        \toprule
        \textbf{Input \textit{S}} & I am pleased with the fast log on, speedy WiFi connection and the long battery life (>6hrs). \\\midrule
        \textbf{RE-prefix} & ``Definition: Solve a relation extraction (RE) task. Given the context, output the most precise semantic relation between the entities 'log on' and 'WiFi connection'. In cases where there is no relationship the output should be NONE. Reason the answer step-by-step. Context: I am pleased with the fast log on, speedy WiFi connection and the long battery life (>6hrs).'' \\\midrule
        \textbf{NER-prefix} & ``Definition: Given the following context, output the relevant entities in it. Reason the answer step-by-step. Context: I am pleased with the fast log on, speedy WiFi connection and the long battery life (>6hrs).'' \\\midrule
        \textbf{Noise-prefix} & ``Definition: elegantly messier nordin fulke wantonness defile sills newland sbu lena hoff nubia cobblestones caddis disliking gaster domicil martialed sylvestre chagall enquires delphic haring niobe intrusive mnes scolex counterpoise detoxification tanglewood sedgwick vintner anker northfield thrilled transvestite echeverria radula lengths abdullah kiri unhinged minefields cloaked restrictive humored refractometer troy cargoes cordate'' \\\bottomrule
    \end{tabular}}
    \caption{Illustration of three prefix types for an input sentence with two aspects (\textit{log on} and \textit{WiFi connection}).}
    \label{tab:prefix-example}
\end{table}

\begin{table}[]
    \centering
    \resizebox{\columnwidth}{!}{
    \begin{tabular}{p{0.2\linewidth} p{0.75\linewidth}}
        \toprule
        \textbf{Input} & Food is always fresh and hot- ready to eat! \\ \midrule
        \textbf{RE-prefix} & -\\ \midrule
        \textbf{NER-prefix} & ``Definition: Given the following context, output the relevant entities in it. Reason the answer step-by-step. Context: Food is always fresh and hot- ready to eat!'' \\\midrule
        \textbf{Noise-prefix} & ``Definition: longmans propulsive kirchen cofactor encoders granitic description carlist yorick accosted outgoings flathead metallization ings surrounds cunliffe relevant quagmire hacked castellana extenders railwaymen windbreak stichting sepia stg jewess bashfulness engrossing fiberboard passionless deb vicente hilbert firft independently inconvenient bloodhound complexed eglantine ricardo casts kebir exoneration undernourishment kerygma extenuate englishmen porridge legitimize'' \\ \bottomrule
    \end{tabular}}
    \caption{Illustration of three prefix types for an input sentence with one aspect (\textit{food}).}
    \label{tab:prefix-example2}
\end{table}

In Tables~\ref{tab:appate}--\ref{tab:appsenti}, we provide details of complete instruction prompts for all five subtasks. Task definition and in-context examples in ATE, ATSC and AOOE subtasks are from \cite{scaria2023instructabsa}. 

\begin{table*}[]
    \centering
    \begin{tabular}{{p{13cm}}}
        \toprule
``\textbf{Definition:} Given the following context, output the relevant entities in it. Reason the answer step-by-step. \\
            Context: I recommend this place to everyone.\\
Afterwards solve the following task\\
\textbf{Definition:} The output will be the aspects (both implicit and explicit) which have an associated opinion that are extracted from the input text. In cases where there are no aspects the output should be noaspectterm.\\
        \textbf{Positive example 1-}\\
        input: With the great variety on the menu , I eat here often and never get bored.\\
        output: menu\\
        \textbf{Positive example 2-}\\ 
        input: Great food, good size menu, great service and an unpretensious setting.\\
        output: food, menu, service, setting\\
        \textbf{Negative example 1-}\\
        input: They did not have mayonnaise, forgot our toast, left out ingredients (ie cheese in an omelet), below hot temperatures and the bacon was so over cooked it crumbled on the plate when you touched it.\\
        output: toast, mayonnaise, bacon, ingredients, plate\\
        \textbf{Negative example 2-}\\
        input: The seats are uncomfortable if you are sitting against the wall on wooden benches.\\
        output: seats\\
        \textbf{Neutral example 1-}\\
        input: I asked for seltzer with lime, no ice.\\
        output: seltzer with lime\\
        \textbf{Neutral example 2-}\\
        input: They wouldnt even let me finish my glass of wine before offering another.\\
        output: glass of wine\\
        \textbf{Now complete the following example-}\\
        input: I recommend this place to everyone. \\
        output: ''\\
        \bottomrule
    \end{tabular}
    \caption{Illustration of an input prompt with NER-prefix for ATE subtask. Words in \textbf{boldface} to ease visualization. }
    \label{tab:appate}
\end{table*}

\begin{table*}[]
    \centering
    \begin{tabular}{{p{13cm}}}
        \toprule
``\textbf{Definition:} Given the following context, output the relevant entities in it. Reason the answer step-by-step.\\
            Context: Boot time is super fast, around anywhere from 35 seconds to 1 minute.\\
Afterwards solve the following task\\
\textbf{Definition:} The output will be 'positive' if the aspect identified in the sentence contains a positive sentiment. If the sentiment of the identified aspect in the input is negative the answer will be 'negative'. \\
        Otherwise, the output should be 'neutral'. For aspects which are classified as noaspectterm, the sentiment is none.\\
        \textbf{Positive example 1-}\\
        input: I charge it at night and skip taking the cord with me because of the good battery life. The aspect is battery life.\\
        output: positive\\
        \textbf{Positive example 2-} \\
        input: Easy to start up and does not overheat as much as other laptops. The aspect is start up.\\
        output: positive\\
        \textbf{Negative example 1-}\\
        input: Also kinda loud when the fan was running. The aspect is fan.\\
        output: negative\\
        \textbf{Negative example 2-}\\
        input: but now i have realized its a problem with this brand. The aspect is brand.\\
        output: negative\\
        \textbf{Neutral example 1-}\\
        input: I took it back for an Asus and same thing, it required me to remove the battery to reset. The aspect is battery.\\
        output: neutral\\
        \textbf{Neutral example 2-}\\
        input: I can always buy and install a camera. The aspect is camera.\\
        output: neutral\\
        \textbf{Now complete the following example-}\\
        input: Boot time is super fast, around anywhere from 35 seconds to 1 minute. The aspect is Boot time.
output: ''\\
        \bottomrule
    \end{tabular}
    \caption{Illustration of an input prompt with NER-prefix for ATSC subtask. Words in \textbf{boldface} to ease visualization.}
    \label{tab:appatsc}
\end{table*}

\begin{table*}[]
    \centering
    \begin{tabular}{{p{13cm}}}
        \toprule
``\textbf{Definition:} Solve a relation extraction (RE) task. Given the context, output the most precise semantic relation between the entities 'spicy tuna roll' and 'asian salad'. In cases where there is no relationship the output should be NONE. Reason the answer step-by-step.\\
        Context: BEST spicy tuna roll , great asian salad .\\
Afterwards solve the following task\\
\textbf{Definition:} The output will be the opinion/describing word of the aspect terms in the sentence. In cases where there are no aspects the output should be none.\\
        \textbf{Positive example 1-}\\
        input: I charge it at night and skip taking the cord with me because of the good battery life . The aspect is battery life.\\
        output: good\\
        \textbf{Positive example 2-}\\
        input: it is of high quality , has a killer GUI , is extremely stable , is highly expandable , is bundled with lots of very good applications , is easy to use , and is absolutely gorgeous. The aspect is GUI.\\
        output: killer\\
        \textbf{Negative example 1-}\\
        input: One night I turned the freaking thing off after using it , the next day I turn it on , no GUI , screen all dark , power light steady , hard drive light steady and not flashing as it usually does . The aspect is GUI.\\
        output: no\\
        \textbf{Negative example 2-}\\
        input: I can barely use any usb devices because they will not stay connected properly . The aspect is usb devices.\\
        output: not stay connected properly\\
        \textbf{Neutral example 1-}\\
        input: However , the multi-touch gestures and large tracking area make having an external mouse unnecessary ( unless you 're gaming ) . The aspect is external mouse.\\
        output: unnecessary\\
        \textbf{Neutral example 2-}\\
        input: I wanted to purchase the extended warranty and they refused , because they knew it was trouble . The aspect is extended warranty.\\
        output: refused\\
        \textbf{Now complete the following example-}\\
        input: BEST spicy tuna roll , great asian salad . The aspect is spicy tuna roll.\\
output: ''\\
        \bottomrule
    \end{tabular}
    \caption{Illustration of an input prompt with RE-prefix for AOOE subtask. Words in \textbf{boldface} to ease visualization.}
    \label{tab:appaooe}
\end{table*}

\begin{table*}[]
    \centering
    \begin{tabular}{p{13cm}}
        \toprule
``\textbf{Definition:} Solve a relation extraction (RE) task. Given the context, output the most precise semantic relation between the entities `brain disease' and `neurotrophic factor'. In cases where there is no relationship the output should be NONE. Reason the answer step-by-step.\\
        Context: The loss of neurotrophic factors such BDNF and CNTF may be associated with the pathogenesis of brain diseases (Chauhan, Siegel, \& Lee, 2001; Jeon et al., 2015; Jeong et al., 2015; Phillips et al., 1991; Sopova, Gatsiou, Stellos, \& Laske, 2014)\\
Afterwards solve the following task\\
\textbf{Definition:} The output will be 'positive' if the aspects identified in the sentence express a positive sentiment for human's health.\\
          If the sentiment of the identified aspects in the input is negative for human's health the answer will be 'negative'. \\
        Otherwise, the output should be 'neutral'.\\
        \textbf{Positive example 1-}\\
        input: Clonidine and tizanidine have been used in the treatment of chronic pain disorders. The aspects are tizanidine and chronic pain.\\ 
        output: positive\\
        \textbf{Positive example 2-}\\ 
        input: One of the recommended standard treatments for diabetic neuropathy is pregabalin. The aspects are pregabalin and diabetic neuropathy.\\
        output: positive\\
        \textbf{Negative example 1-}\\
        input: NFAT promotes carcinoma invasive migration through glypican-6. The aspects are carcinoma and glypican.\\
        output: negative\\
        \textbf{Negative example 2-}\\
        input: frutescens on streptozotocin induced diabetes in neonatal (n-STZ)-type II Wistar rats. The aspects are streptozotocin and diabetes.\\
        output: negative\\
        \textbf{Neutral example 1-}\\
        input: He had not noted any insect bites and denied ingesting any raw meat, freshwater fish, or liver. The aspects are liver and insect bite.\\
        output: neutral\\
        \textbf{Neutral example 2-}\\
        input: Non-heme iron concentrations in cooked sausages are shown in Figure 1. The aspects are sausages and heme.\\
        output: neutral\\
        \textbf{Now complete the following example-}\\
        input: The loss of neurotrophic factors such BDNF and CNTF may be associated with the pathogenesis of brain diseases (Chauhan, Siegel, \& Lee, 2001; Jeon et al., 2015; Jeong et al., 2015; Phillips et al., 1991; Sopova, Gatsiou, Stellos, \& Laske, 2014) The aspects are brain disease and neurotrophic factor.\\
output: ''
       \\ \bottomrule
    \end{tabular}
    \caption{Illustration of an input prompt with RE-prefix for ERSA subtask. Words in \textbf{boldface} to ease visualization.}
    \label{tab:appersa}
\end{table*}

\begin{table*}[]
    \centering
    \begin{tabular}{p{13cm}}
        \toprule
``\textbf{Definition:} teases faunas aap pills esophagus overplus bhp inches diplococcus thatcher substituting sama lancey cortex directness pastoralism illicitly urbain inserting educated belonging unde fragmented pepperell scoria chee altogether purifications purchase prostatic neuralgia sternum germinate maneuvering holier gabble conjuration rac dolley hartshorn illam postgate macdonnell tripoli pitot massa radford porcine iiib laufer\\
Afterwards solve the following task\\
\textbf{Definition:} Solve the following task. The output will be 'positive' if the identified aspect of a given entity in the input sentence contains a positive sentiment. If the sentiment of the identified aspect in the input is negative the answer will be 'negative'. \\
        Otherwise, the output should be 'neutral'.\\
        \textbf{Positive example 1-}\\
        input: Of course LOCATION1 is also very central. The entity is LOCATION1, the aspect is transit-location.\\
        output: positive\\
        \textbf{Positive example 2-} \\
        input: If I were you I would look nearby LOCATION1. The entity is LOCATION1, the aspect is general.\\
        output: positive\\
        Positive example 3-\\
        input: LOCATION1 is an ugly cold place but it isn't dangerous. The entity is LOCATION1, the aspect is safety.\\
        output: positive\\
        \textbf{Negative example 1-} \\
        input: I'd stay away from LOCATION1. The entity is LOCATION1, the aspect is general.\\
        output: negative\\
        \textbf{Negative example 2-}\\
        input: LOCATION1 is a nice area, but apartments are very pricey. The entity is LOCATION1, the aspect is price.\\
        output: negative\\
        \textbf{Negative example 3-}\\
        input: LOCATION1 is all junkies. The entity is LOCATION1, the aspect is safety.\\
        output: negative\\
        \textbf{Now complete the following example-}\\
        input: LOCATION1 is in Greater London  and is a very safe place. The entity is LOCATION1, the aspect is safety.\\
output: ''\\
        \bottomrule
    \end{tabular}
    \caption{Illustration of an input prompt with Noise-prefix for SentiHood (ATSC) subtask. Words in \textbf{boldface} to ease visualization.}
    \label{tab:appsenti}
\end{table*}

\end{document}